\documentclass[runningheads]{llncs}
\usepackage[T1]{fontenc}
\usepackage{graphicx}
\usepackage{subfig}
\usepackage{hyperref}

\topskip
\parskip

\usepackage{courier} 
\usepackage{listings, xcolor}

\usepackage[many]{tcolorbox}
\usepackage{xcolor}
\usepackage{varwidth}
\usepackage{environ}
\usepackage{xparse}

\colorlet{punct}{red!60!black}
\definecolor{background}{HTML}{FFFFFF}
\definecolor{delim}{RGB}{20,105,176}
\colorlet{numb}{magenta!60!black}

\lstdefinelanguage{json}{
    basicstyle=\footnotesize\ttfamily,
    numbers=left,
    numberstyle = \tiny,
    showstringspaces=false,
    breaklines=true,
    tabsize = 2, 
    literate=
     *{0}{{{\color{numb}0}}}{1}
      {1}{{{\color{numb}1}}}{1}
      {2}{{{\color{numb}2}}}{1}
      {3}{{{\color{numb}3}}}{1}
      {4}{{{\color{numb}4}}}{1}
      {5}{{{\color{numb}5}}}{1}
      {6}{{{\color{numb}6}}}{1}
      {7}{{{\color{numb}7}}}{1}
      {8}{{{\color{numb}8}}}{1}
      {9}{{{\color{numb}9}}}{1}
      {:}{{{\color{punct}{:}}}}{1}
      {,}{{{\color{punct}{,}}}}{1}
      {\{}{{{\color{delim}{\{}}}}{1}
      {\}}{{{\color{delim}{\}}}}}{1}
      {[}{{{\color{delim}{[}}}}{1}
      {]}{{{\color{delim}{]}}}}{1},
}

\newlength{\bubblewidth}
\AtBeginDocument{\setlength{\bubblewidth}{0.9\textwidth}} 
\definecolor{bubbleprimary}{RGB}{59,113,202}
\definecolor{bubblelight}{RGB}{159,166,178}
\definecolor{bubbledark}{RGB}{51,45,45}

\newcommand{\bubble}[4]{
  \tcbox[
    colback=#1,
    colframe=#1,
    #2,
    top=0pt,    
    bottom=0pt, 
    left=0pt,   
    right=0pt,  
  ]{\color{#3}\begin{varwidth}{\bubblewidth}#4\end{varwidth}}
}

\ExplSyntaxOn
\seq_new:N \l__ooker_bubbles_seq
\tl_new:N \l__ooker_bubbles_first_tl
\tl_new:N \l__ooker_bubbles_last_tl

\newcommand{\bubblefontsize}{\scriptsize} 

\NewEnviron{rightbubbles}
 {
  \raggedleft\bubblefontsize\sffamily
  \seq_set_split:NnV \l__ooker_bubbles_seq { \par } \BODY
  \int_compare:nTF { \seq_count:N \l__ooker_bubbles_seq < 2 }
   {
    \bubble{bubbleprimary}{rounded~corners}{white}{\BODY}
   }
   {
    \seq_pop_left:NN \l__ooker_bubbles_seq \l__ooker_bubbles_first_tl
    \seq_pop_right:NN \l__ooker_bubbles_seq \l__ooker_bubbles_last_tl
    \bubble{bubbleprimary}{sharp~corners=southeast}{white}{\l__ooker_bubbles_first_tl}\par
    \seq_map_inline:Nn \l__ooker_bubbles_seq
     {
      \bubble{bubbleprimary}{sharp~corners=east}{white}{##1}\par
     }
    \bubble{bubbleprimary}{sharp~corners=northeast}{white}{\l__ooker_bubbles_last_tl}\par
   }
 }
\NewEnviron{leftbubbles}
 {
  \raggedright\bubblefontsize\sffamily
  \seq_set_split:NnV \l__ooker_bubbles_seq { \par } \BODY
  \int_compare:nTF { \seq_count:N \l__ooker_bubbles_seq < 2 }
   {
    \bubble{bubblelight}{rounded~corners}{black}{\BODY}
   }
   {
    \seq_pop_left:NN \l__ooker_bubbles_seq \l__ooker_bubbles_first_tl
    \seq_pop_right:NN \l__ooker_bubbles_seq \l__ooker_bubbles_last_tl
    \bubble{bubblelight}{sharp~corners=southwest}{black}{\l__ooker_bubbles_first_tl}\par
    \seq_map_inline:Nn \l__ooker_bubbles_seq
     {
      \bubble{bubblelight}{sharp~corners=west}{black}{##1}\par
     }
    \bubble{bubblelight}{sharp~corners=northwest}{black}{\l__ooker_bubbles_last_tl}\par
   }
 }
 \NewEnviron{leftbubblessecondary}
 {
  \raggedright\bubblefontsize\sffamily
  \seq_set_split:NnV \l__ooker_bubbles_seq { \par } \BODY
  \int_compare:nTF { \seq_count:N \l__ooker_bubbles_seq < 2 }
   {
    \bubble{bubbledark}{rounded~corners}{white}{\BODY}
   }
   {
    \seq_pop_left:NN \l__ooker_bubbles_seq \l__ooker_bubbles_first_tl
    \seq_pop_right:NN \l__ooker_bubbles_seq \l__ooker_bubbles_last_tl
    \bubble{bubbledark}{sharp~corners=southwest}{white}{\l__ooker_bubbles_first_tl}\par
    \seq_map_inline:Nn \l__ooker_bubbles_seq
     {
      \bubble{bubbledark}{sharp~corners=west}{white}{##1}\par
     }
    \bubble{bubbledark}{sharp~corners=northwest}{white}{\l__ooker_bubbles_last_tl}\par
   }
 }
\ExplSyntaxOff

\begin{document}

\title{PROMISE: A Framework for Developing Complex Conversational Interactions (Technical Report)}
\titlerunning{Framework for Conversational Interactions}

\author{Wenyuan~Wu\inst{1} \and
Jasmin~Heierli\inst{2} \and
Max~Meisterhans\inst{2} \and
Adrian~Moser\inst{2} \and
Andri~Färber\inst{2} \and
Mateusz~Dolata\inst{1} \and
Elena~Gavagnin\inst{2} \and
Alexandre~de Spindler\inst{2} \and
Gerhard~Schwabe\inst{1}}

\authorrunning{W. Wu et al.}

\institute{University of Zurich, Zürich, Switzerland
\email{\{wenyuan,dolata,schwabe\}@ifi.uzh.ch}
\and Zurich University of Applied Sciences, Winterthur, Switzerland
\email{\{heej,meix,mosa,faer,gava,desa\}@zhaw.ch}}

\maketitle

\begin{abstract}
The advent of increasingly powerful language models has raised expectations for language-based interactions. However, controlling these models is a challenge, emphasizing the need to be able to investigate the feasibility and value of their application.
We present PROMISE\footnote{Available at: \url{https://github.com/zhaw-iwi/promise}}, a framework that facilitates the development of complex language-based interactions with information systems. Its use of state machine modeling concepts enables model-driven, dynamic prompt orchestration across hierarchically nested states and transitions. This improves the control of the behavior of language models and thus enables their effective and efficient use.

In this technical report we show the benefits of PROMISE in the context of application scenarios within health information systems and demonstrate its ability to handle complex interactions. We also include code examples and present default user interfaces available as part of PROMISE.
\keywords{Framework \and Prompt Orchestration \and Language Models}
\end{abstract}


\section{Introduction}
\label{sec:introduction}
Natural language-based interactions enhance the use of information systems beyond areas where traditional inputs such as keyboards and touch are impractical. For example, voice assistants such as Alexa, Google Assistant, and Siri~\cite{kepuska_next-generation_2018} have made substantial progress in question answering and command execution. More advanced services like GitHub Copilot and Microsoft 365 Copilot integrate language processing with complex data types, including source code and office documents, to provide sophisticated, context-aware functionality. These systems demonstrate a shift towards more adaptive and personalized system value through complex language-based interactions.

Language-based interactions are also gaining importance within health information systems~\cite{SongPerformanceLLMinHealth,PengStudyLLM4Med}, showing promise in achieving key medical objectives. These include achieving patient-centered medicine by adapting to patients' needs~\cite{LanbergPatientCeterdnessConcept}, better handling of text data in electronic health records (EHR)~\cite{HossainNLP4EHR}, and streamlining physician-patient collaboration, leading to more targeted and effective consultations and treatments~\cite{FarberClosingTheLoop}.

Patient treatment has been shown to benefit from medical consultations where treatment goals are discussed, patient concerns are addressed and the physician-patient relationship is strengthened by compassionate communication~\cite{stephensComplexConversationsHealthcare2021}. This entails complex interactions involving several successive or alternating conversation threads, possibly leading to nested conversations. Consequently, if a system is to have a similar treatment benefit between consultations, it must be able to handle such complex interactions.

While the recent advent of more powerful language models (LM) promises novel and advanced language-based interactions, more progress must be made in supporting LM application development~\cite{HadiLLMSurveyChallenges}. Training LMs from scratch to serve a specific purpose is resource-intensive and often impractical for typical development projects. Although fine-tuning can tailor LM responses, it demands meticulous data preparation, making fast, iterative experimentation difficult. In contrast, prompt engineering allows to bypass traditional pre-training and fine-tuning bottlenecks. However, the specification of a complex interaction entails complex prompts, which lack reliability. Ultimately, neither approach fully addresses the challenges arising when complex interactions are designed, integrated with information systems, implemented in variants and improved iteratively.

In this paper, we present PROMISE (Prompt-Orchestrating Model-driven Interaction State Engineering), an application development framework addressing the need for more support when designing and implementing complex language-based interactions. PROMISE bridges the gap between the requirements of such interactions and the use of language models to enable them. Framework support is based on a model that can capture a wide range of requirements and effectively control the use of LMs while leveraging their full capabilities.

In the following Sect.\ref{sec:relatedWork}, existing approaches to developing language-based interactions are reviewed. Sect.\ref{sec:frameworkModel} introduces PROMISE, with Sect.\ref{sec:frameworkApplication} presenting its application and Sect.\ref{sec:frameworkImplementation} detailing its design and implementation. The capabilities of PROMISE are validated through an advanced application in Sect.\ref{sec:validation}, concluding with remarks in Sect.\ref{sec:conclusion}.
\section{Related Work}
\label{sec:relatedWork}

LMs offer increased flexibility for open-ended conversations, yet directing their behavior remains challenging~\cite{HadiLLMSurveyChallenges}. Leveraging LMs' zero- and few-shot learning abilities, \textit{Prompt Engineering} provides an efficient method for behavior control~\cite{korzynski_artificial_2023,white_prompt_2023}. While many prompt articulation strategies were developed~\cite{wei_chain--thought_2022,fernando2023promptbreeder,chuSurveyChainThought2023}, this alone cannot ensure consistent LM behavior in complex interactions.

Overly detailed prompts that cover the entire interaction risk confusing sequences or levels, and overly broad prompts risk missing expected responses, may induce baseless \textit{hallucinations}, or even introduce vulnerabilities~\cite{mozes2023use}. PROMISE, therefore, follows the idea of segmenting complex interactions into a sequence of more specific tasks. This was shown to enhance control and predictability while harnessing LM capabilities~\cite{helland2023divide}.

The emergence of prompting techniques has brought readily available frameworks supporting the development of LM applications with implicit prompting support. For example, LangChain~\cite{LangChainGitHub} provides seamless integrations with other system components such as retrieval-augmented generation (RAG), optionally involving multiple data sources. However, LangChain's support for nested interactions is limited because it cannot select distinct and potentially nested segments and detect specific characteristics. In contrast, PROMISE offers explicit segment selection, detection of characteristics, and the means to map their occurrence to specified actions and follow-up interactions.

Beyond these practical efforts to enhance LM controllability and augmentation, there has been conceptual progress in supporting complex interactions, particularly in dialogue management research~\cite{BrabraDialogueMgmtSurvey}. Although much of this prior work was not focused on controlling LMs through prompts, they provide solutions that can be adapted for LM applications. For example, IRIS~\cite{fastIrisConversationalAgent2018} allows users to select commands and engage in nested interactions for argument completions, using state machines with shared memory for seamless information flow.

Our approach introduces a semantically richer state model where states can include other states, allowing for more comprehensive interaction specifications to be achieved through state machines alone. We extend this model to attach partial prompts to both states and transitions, facilitating dynamic prompt composition and augmentation. This extension enables a broader range of interaction architectures to leverage LM capabilities more effectively.

By using state machines to specify interactions, PROMISE enables the design and reuse of abstractions and patterns in compositional conversational flows~\cite{BougueliaCompositionalConversationalFlows}. This includes interactions involving slot values and nested intents, and integrating external services for decision-making. However, PROMISE enables interactions with more complex nesting structures beyond handling of slot values, such as for patient health inquiries or educational coaching sessions.

\section{PROMISE Interaction Model}
\label{sec:frameworkModel}

The PROMISE framework simplifies the development of complex language-based interactions, minimizing both design and coding efforts through state machine modeling. This approach enhances the application and control of language models (LMs). We illustrate this with health domain examples, showcasing how state models shape the design and implementation of conversational behaviors.

\begin{figure}[htb]
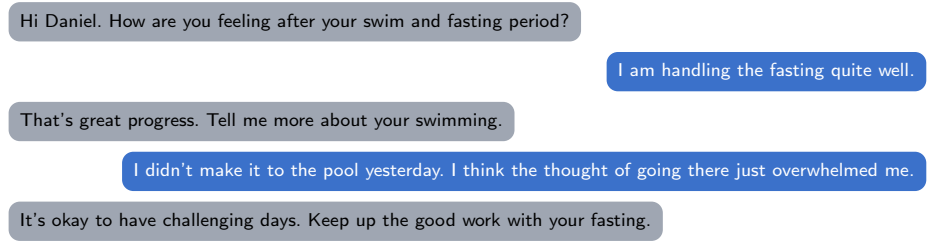

    \begin{leftbubbles}
    Hi Daniel. How are you feeling after your swim and fasting period?
    \end{leftbubbles}
    \begin{rightbubbles}
    I am handling the fasting quite well.
    \end{rightbubbles}
    \begin{leftbubbles}
    That's great progress. Tell me more about your swimming.
    \end{leftbubbles}
    \begin{rightbubbles}
    I didn't make it to the pool yesterday. I think the thought of going there just overwhelmed me.
    \end{rightbubbles}
    \begin{leftbubbles}
    It's okay to have challenging days. Keep up the good work with your fasting.
    \end{leftbubbles}
    \caption{Single-State Interaction: System (Gray) \& User (Blue)}
    \label{con:singleStateConversation}
\end{figure}

The conversation in Fig.~\ref{con:singleStateConversation} is a daily check-in interaction with patients using a health information system. Such interactions aim to to assess their well-being related to their chronic condition and therapy plan.

PROMISE supports including patient data from their electronic health record (EHR). As a result, the system can have a personalized conversation, e.g., by addressing the patient by their name and asking about their perceived progress regarding their therapy activities. The conversational behavior is designed to use open-ended questions and empathetic dialogue to create a supportive environment. Furthermore, PROMISE provides a means of ending an interaction based on defined conditions arising during the conversation. In this case, the interaction is designed to end when there is enough information about the patient's adherence and well-being. Finally, the end of the interaction is configured to summarize the conversation in a specific form. In this example, the following summary is generated for a dashboard component that informs the attending physician in a follow-up consultation.

\begin{lstlisting}[language=json, numbers=none]
{"adherence": "Partial; fasted successfully, missed swimming.",
 "wellbeing": "Fasting is going well; feelings of being overwhelmed impacted swimming attendance."}
\end{lstlisting}

\begin{figure}[htb]
    \centering
    \includegraphics[scale=0.6]{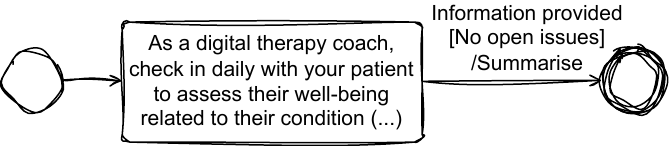}
    \caption{Single-State Model}
    \label{fig:singleStateModel}
\end{figure}

With PROMISE, the state machine shown in Fig.~\ref{fig:singleStateModel} is used to model this interaction. The state machine consists of a single state in the middle, an initial node on the left, and a final node on the right. Transitions lead from the initial node to the state, and from the state to the final node. Transitions depend on triggers and guards, and actions are executed when followed.

The state is annotated with the \textit{state prompt} "As a digital therapy coach, \dots" which will be used to control the LM while the interaction is in that state. The outgoing transition that leads to the final node is annotated with prompts indicated by "Information provided", "No open issues", and "Summarise". These prompts control the LM when evaluating the conversation concerning transition triggers, guards, and actions. As shown next, PROMISE transparently composes more complex prompts from simple prompts attached to states and transitions.

When the interaction is started, the initial node is used to identify its first state. In this state, PROMISE will create a $Prompt_{Composed}$ by concatenating the state prompt
\begin{equation*}
    Prompt_{State} = \text{"As a digital therapy coach, \ldots"}
\end{equation*}
with the \textit{state starter prompt}
\begin{equation*}
    Prompt_{State Starter} = \text{"\ldots compose a single, very short message \ldots}
\end{equation*}
such as to obtain the composed prompt
\begin{align*}
    Prompt_{Composed}  &= Prompt_{State} + Prompt_{Starter} \\
                    &= \text{"As a digital \ldots compose a single \ldots"}
\end{align*}
The state starter prompt is an optional extension of the state prompt to be used if the state is configured to start the conversation, as is the case in this example. Any prompt can be enriched with placeholders that are filled at runtime. In our example, the state prompt "As a digital therapy coach, \dots. Meet \{patient\}." contains the placeholder \texttt{\{patient\}} that will be replaced with patient data obtained from the EHR at runtime. 

The composed prompt is then used to instruct the LM. In the example conversation above, the LM completion returned the utterance 
\begin{equation*}
    Utterance_{Starter} = \text{"Hi Daniel. How are you feeling today after your \ldots"}
\end{equation*}
which opens the conversation with the patient.

For every utterance from the patient, such as the first utterance, "I’m doing quite well \dots", all outgoing transitions are checked before the LM is used to generate a response to the patient. For this purpose, the conversation held within that state so far, referred to as $Utterances^t_{State}$, is extended with the incoming user utterance $Utterance_{User}$ as follows.
\begin{align*}
    Utterances^{t+1}_{State}    &= Utterances^t_{State} + Utterance_{User} \\
                                &= [\text{"Hi Daniel. How are you \ldots"}, \text{"I’m doing quite well \ldots"}]
\end{align*}
To check a transition, the decision prompts $Prompt_{Trigger}$, $Prompt_{Guard}$, and action prompt $Prompt_{Action}$ --- which all may contain placeholders or consist of code instead of prompts --- are used to obtain decisions from the LM about whether or not the transition should be followed or execute an action if it is followed. For example, in the case of a transition trigger, the composed prompt
\begin{align*}
    Prompt_{Comp.}  &= Prompt_{Trigger} + Utterances^{t+1}_{State} \\
                    &= \text{"Examine the conversation\ldots, decide if\ldots patient provided\ldots"} \\
                    &+ [\text{"Hi Daniel. How are you \ldots"}, \text{"I’m doing quite well \ldots"}] 
\end{align*}
is created to let the LM decide whether the conversation so far contains indications that the patient provided the expected information about their adherence and well-being. While the first patient response talks about fasting, no swimming information has been mentioned so far. Consequently, this transition trigger does not pass, and the interaction stays in the current state.

Multiple decisions may be attached to a single transition. Each decision may contain a prompt for LM-based evaluation or code that implements any other evaluation. In our example, a second decision serves as a transition guard, and instructs the LM to decide whether there are no open issues mentioned by the patient that should prevent the current interaction from being concluded unexpectedly.

When the interaction stays in the current state, the state prompt and utterances collected so far are included in the newly composed prompt
\begin{align*}
    Prompt_{Composed}  &= Prompt_{State} + Utterances^{t+1}_{State} \\
                    &= \text{"As a digital \ldots"} \\
                    &+ [\text{"Hi Daniel. How are you \ldots"}, \text{"I’m doing quite well \ldots"}] 
\end{align*}
which is used to obtain the subsequent response to the patient from the LM. These responses are also appended to the state utterances. As seen in the example interaction above, the conversation stays in the same state as long as the expected information is incomplete. When all the information is provided, and the transition decisions pass, the conversation continues with the subsequent state attached to the transition. In this example, the subsequent state is a final state, where the interaction ends with a final message to the patient.

Transitions may include multiple actions that contain a prompt or code. In our example, the action is a prompt with which the composed prompt
\begin{align*}
    Prompt_{Composed}  &= Prompt_{Action} + Utterances_{State} \\
                    &= \text{"Summarize the conversation \ldots for the physician's review."} \\
                    &+ [\text{"Hi Daniel. How \ldots"}, \ldots, \text{"\ldots again tomorrow."}]
\end{align*}
is created and used to instruct the LM to summarise the information provided by the patient for the attending physician's review. The generated summary is stored in interaction storage and thus made available either to any other state, transition decision or action, or to any other information system component, such as the physician's dashboard.

\begin{figure}[htb]
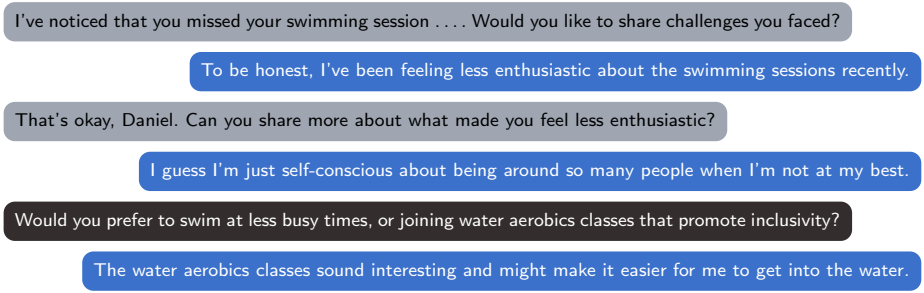

    \begin{leftbubbles}
    I've noticed that you missed your swimming session \dots. Would you like to share challenges you faced?
    \end{leftbubbles}
    \begin{rightbubbles}
    To be honest, I've been feeling less enthusiastic about the swimming sessions recently.
    \end{rightbubbles}
    \begin{leftbubbles}
    That's okay, Daniel. Can you share more about what made you feel less enthusiastic?
    \end{leftbubbles}
    \begin{rightbubbles}
    I guess I'm just self-conscious about being around so many people when I'm not at my best.
    \end{rightbubbles}
    \begin{leftbubblessecondary}
    Would you prefer to swim at less busy times, or joining water aerobics classes that promote inclusivity?
    \end{leftbubblessecondary}
    \begin{rightbubbles}
    The water aerobics classes sound interesting and might make it easier for me to get into the water.
    \end{rightbubbles}

    \caption{Multi-State Interaction: First (Light Gray) \& Second (Dark) State}
    \label{con:multiStateConversation}
\end{figure}

Language-based interactions often require a sequence of goals to be met, as opposed to the previous single state example, where the only goal was to assess patient adherence and well-being. To illustrate PROMISE's capability to support more complex interactions, we introduce the example interaction in Fig.~\ref{con:multiStateConversation} that follows up on the daily check-in presented before. This interaction was initiated since the patient mentioned missing a therapy activity (swimming). It first aims to obtain a reason for the patient's omission (light grey) and then to propose ways to adapt this activity in order to increase the patient's adherence to this activity (dark grey).

\begin{figure}[htb]
    \centering
    \includegraphics[width=\textwidth]{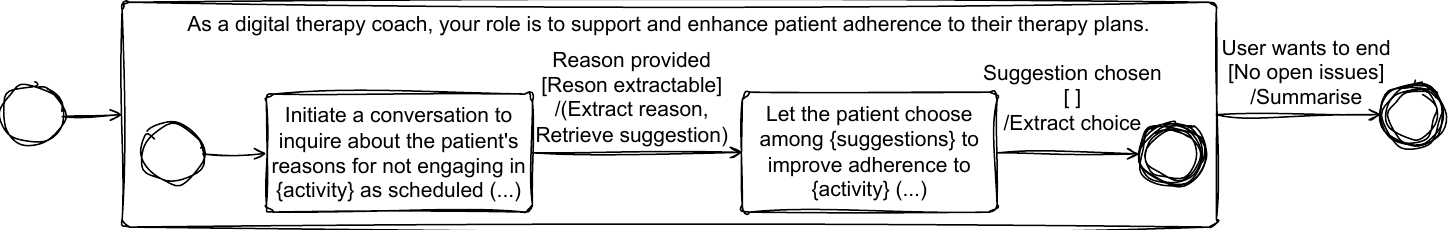}
    \caption{Multi-State Model}
    \label{fig:multiStateModel}
\end{figure}

Figure~\ref{fig:multiStateModel} shows the state machine modeling this conversational behavior. It includes three novelties compared to the previous single-state interaction: 1) there are multiple states following up on each other. 2) special-purpose states are involved, such as patient assessments and choices. 3) there is an outer state containing the sequence of special-purpose states.

First, the ability to create a sequence of states follows from the ability to use transitions. Any state may have arbitrarily many outgoing transitions --- each triggered and guarded by decisions and accompanied by actions --- that each point to another state, thus supporting different interaction flows. While this effectively supports the creation of a directed and cyclic graph of states, we stick to a simple sequence of states in this example. The interaction starts with a state assessing the patient's reasoning for missing the swimming activity. When the patient gives a sufficient reason, the interaction transitions to the next state in which options are presented. Once the patient has chosen one of these options, the interaction ends in a final node.

Second, PROMISE comes with predefined special-purpose states that address recurring requirements. For example, an \textit{Activity Gap Inquiry State} obtains the patient's reasons for missing the activity. To instantiate such a state, developers simply provide the activity that was missed. Similarly, a \textit{Single Choice State} assumes a list of choices to be made available to the patient. These are two examples from our library of predefined states designed to be reusable in various interaction scenarios. These states have in common that their state prompt and starter prompt and their transitions, including decisions and actions, are generated automatically, effectively reducing development efforts. 

The third novelty relates to PROMISE's ability to support nested conversations by specifying state machines which may behave at different levels --- seemingly simultaneously.  In PROMISE, an outer state follows the complete conversation within all inner states. This is because the outer state maintains its own utterances as a composition of all inner state utterances.
\begin{equation*}
    Utterances^{t}_{Outer State} = Utterances^{t}_{Gather State} + Utterances^{t}_{Single Choice State}
\end{equation*}
Therefore, each transition attached to an outer state reacts to a more extensive interaction segment when making decisions or performing actions. In our example, the patient may indicate they want to pause or stop interacting at any state of the inner interaction. For this purpose, the outer state has its own transition, capturing this patient's wish and leading to a final state. To illustrate this, the composed prompt for such a trigger decision is created as
\begin{equation*}
    Prompt_{Composed}  = Prompt_{Trigger} + Utterances_{Outer State} 
\end{equation*}
and therefore this trigger can detect conversational characteristics manifesting over longer-lasting interaction windows.

Moreover, if an outer state has its own state prompt, this prompt is automatically appended to the state prompts of all its inner states. For example, the composed prompt of any of the inner states shown in Fig.~\ref{fig:multiStateModel} is created as
\begin{equation*}
    Prompt_{Composed}  = Prompt_{Outer State} + Prompt_{Inner State} + Utterances_{Inner State}
\end{equation*}
An outer state thus enables developers to inject partial conversational behaviors that act in the scope of more expansive windows of interactions. In the current example, the role prompt "As a digital therapy coach, \dots" is attached to the outer state, and therefore it does not need to be repeated among all inner states.




\section{PROMISE Application}
\label{sec:frameworkApplication}

This section delves into the practical aspects of creating language-based interactions. We present the code implementing the interactions introduced in the previous Sect.~\ref{sec:frameworkModel}, highlighting the straightforward process facilitated by PROMISE.

An interaction such as the one specified by the state model shown in Fig.~\ref{fig:singleStateModel} is implemented by creating instances of the state model concepts \texttt{State} and \texttt{Transition}. A \texttt{State} is created using a \textit{state prompt}, \textit{state name}, \textit{starter prompt}, and a list of transitions out of this state.
\begin{lstlisting}[language=Java, numbers=none, escapeinside={(*@}{@*)}]
State state = new State("As a therapy coach...", "DailyCheckIn", 
    "... compose a message ...", List.of(transition));
\end{lstlisting}
Since the transitions must be provided, those are created beforehand. The following lines show the creation of a \texttt{Transition} with a trigger, guard, and action.
\begin{lstlisting}[language=Java, numbers=none, escapeinside={(*@}{@*)}]
Storage storage = new Storage();
Decision trigger = new StaticDecision("Examine the conversation 
    ...decide if...patient provided...");
Decision guard = new StaticDecision("Examine the conversation 
    ...confirm that...no open issues...");
Action action = new StaticExtractionAction(
    "Summarize the conversation...", storage, "summary");
Transition transition = new Transition(
    List.of(trigger, guard), List.of(action), new Final());
\end{lstlisting}
The trigger, guard, and action are provided as a list of \texttt{Decision} and a list of \texttt{Action} objects. Transitions may thus be set to depend on multiple decisions and to execute multiple actions. In this example, there is one static decision containing a trigger prompt, another containing a guard prompt, and one static action containing a prompt that summarizes the conversation.

However, PROMISE also addresses other requirements for decisions and actions. For example, it supports dynamic decisions and actions. Those allow data extracted from previous states or other information system components, such as a database or other services, to be injected into prompt template placeholders. As another example, actions may be implemented with code, such as interacting with a database or other services.

The action in this example will summarise the conversation, and a \texttt{Storage} is assigned to this action, where the summary will be put and made available to other system components. PROMISE provides a simple key-value-based storage that may be accessed by any state or transition, to either store or retrieve values, throughout the entire interaction.

The last argument of the transition constructor is the subsequent state that will be attained if the transition is followed. In this single-state interaction, the subsequent state is a \texttt{Final} node where the interaction comes to an end.

Finally, an \texttt{Agent} wraps the state machine and provides the functionality required when integrating the interaction with an information system, as exemplified by the following lines of code.
\begin{lstlisting}[language=Java, numbers=none, escapeinside={(*@}{@*)}]
Agent agent = new Agent(state);
String conversationStarter = agent.start();
String response = agent.respond("I am handling the fasting...");
\end{lstlisting}
The method \texttt{respond()} can be invoked repeatedly for each user utterance. The agent encapsulates the inner workings of the state machine by maintaining a current state and hiding transitions. The \texttt{start()} method may be used to let the system start the conversation. 

In what follows, we focus on the three novelties introduced in the context of the multi-state interaction specified by the state model shown in Fig.~\ref{fig:multiStateModel}. We begin with instantiating special-purpose states made available as part of PROMISE. The following lines of code show the instantiation of two special-purpose states. 
\begin{lstlisting}[language=Java, numbers=none, escapeinside={(*@}{@*)}]
State choice = new SingleChoiceState("Choice", new Final(),
    storage, "suggestionsOffered", "suggestionChosen");
State reason = new ActivityGapInquiryState("Reason", choice,
    storage, "activityMissed", "reasonProvided");
\end{lstlisting}
An \texttt{ActivityGapInquiryState} involves the patient in a conversation to determine why they missed an activity from the therapy plan. In our own project, a system component outside of PROMISE retrieves activity data collected by a smartwatch and compares this data to the therapy plan in the EHR. When a gap is detected, this system component puts a description of the gap into the storage provided by PROMISE and triggers the interaction with the patient. The storage key "activityMissed" thus points to the missed activity, which will be injected into the state prompt. The storage key "reasonProvided" is used to store the reason obtained from the patient. Note that this state encapsulates the creation of a transition triggered by the fact that the patient provided a valid reason, guarded by the extractability of the reason provided, and executing an action that extracts the reason provided and puts it into the storage. The subsequent state named "choice" must be provided and thus created beforehand.

A \texttt{SingelChoiceState} reads a list of options accessed from the storage with the key "suggestionsOffered". Once the patient provides a reason for missing their activity, a system component outside of PROMISE retrieves options for the patient to adapt their therapeutic activity. It puts these into storage before entering this state. The patient is then involved in a conversation in which the available options are presented, which ends once the patient chooses and agrees on one of these options. As with the previous state, all transitions, decisions, and actions are created implicitly. The action extracts the option chosen and puts it into storage with the key "suggestionChosen", where the choice is available for further processing. Following this extraction, this two-state interaction ends in a \texttt{Final} node.

Such special-purpose states reduce coding efforts, as can be observed by comparing the code required for this two-state interaction to the previous single-state interaction. This is because the special-purpose states use predefined prompts, transitions, decisions, and actions.

We now show the implementation of multi-layered interactions utilizing nested states. An \texttt{OuterState} created with the lines of code below contains a pointer to the beginning of the inner state, which, in this example, is the \texttt{Activity\-Gap\-Inquiry\-State} created above.
\begin{lstlisting}[language=Java, numbers=none, escapeinside={(*@}{@*)}]
Transition toFinal = new Transition(List.of(...), List.of(...), new Final());
State outerState = new OuterState(
    "As a digital therapy coach, ...", "Therapy Coach",
    List.of(toFinal), reason);
\end{lstlisting}
This outer state has its own transition that reacts to the patient possibly wishing to end the interaction at any time. Since the outer state maintains its own utterances, including all uttereances of all its inner states, it can react to events occurring at any time or that can only be detected by observing more comprehensive interaction segments.

Furthermore, this outer state specifies its own state prompt, "As a digital therapy coach, ..." which will be inherited by all inner states. Thus, developers can avoid repeating this role prompt across all inner states. Another benefit of such prompt inheritance is that the role prompt can be adapted in one single outer state, while this adaptation is effective for all inner states.

\section{PROMISE Implementations}
\label{sec:frameworkImplementation}

Since the PROMISE framework aims to facilitate feasibility studies and proof-of-value experiments~\cite{NunamakerRigorRelevance}, it is designed to support rapid prototyping. We provide a Python implementation that can be used for in-memory experimentation. State machines can be efficiently implemented and tested within a notebook environment, allowing for interactive experimentation with the resulting interactions.

However, our projects require engaging a diverse group, including patients and physicians, which is often made possible with web-based front-end applications. Given HTTP's stateless nature, this demands back-end database support. Consequently, we offer a Java-based version of PROMISE utilizing Spring Boot\footnote{\url{https://spring.io/projects/spring-boot}} for its state-of-the-art REST and ORM capabilities.

\begin{figure}[htb]
    \centering
    \subfloat[\centering Interaction]{{\includegraphics[width=.247\linewidth]{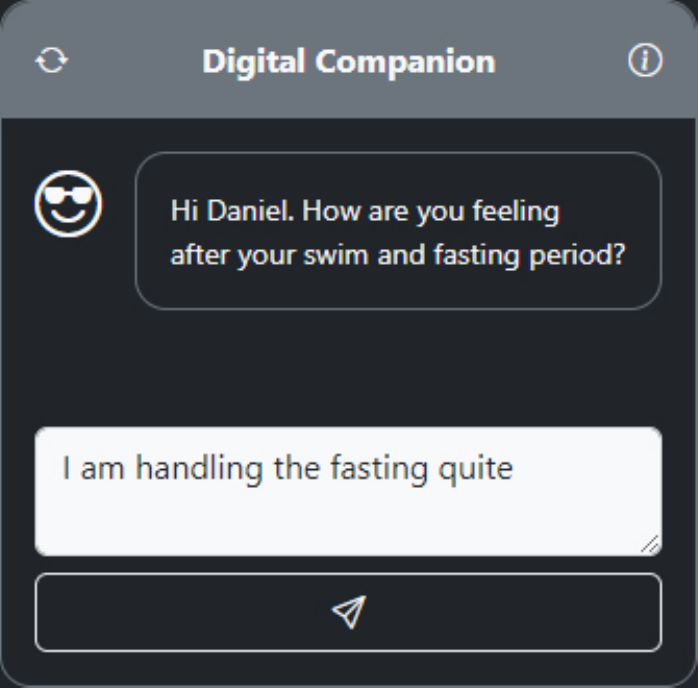}}}%
    \qquad
    \subfloat[\centering Interaction Management]{{\includegraphics[width=.69\linewidth]{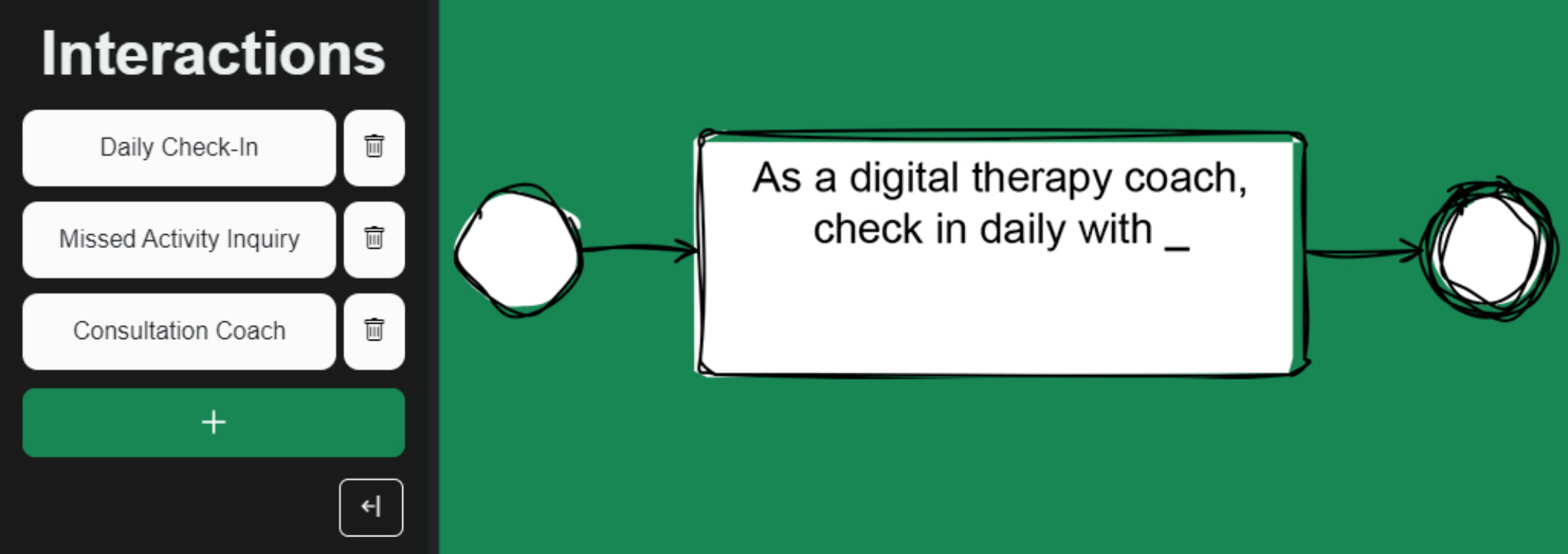}}}%
    \caption{PROMISE Front-Ends (Subset)}
    \label{fig:frontEnds}
\end{figure}

The Java version of PROMISE includes a set of ready-to-use front-ends that support experiments. For example, the left side of Fig.~\ref{fig:frontEnds} shows the front-end that can be used to test state machines with individual users. Once deployed, a unique identifier generated by PROMISE can be appended to the URL handed out to test-users. As a result, each test-user can interact with their own instance, while all interactions can then be evaluated. This front-end is fully responsive and may be adapted to address specific needs.

The other front-end, shown on the right in Fig.~\ref{fig:frontEnds}, supports the management of state machines. The pane to its left provides access to all machines available in the database. The green button opens a tool on the right, where new state machines can be designed and created.

The PROMISE back-end REST API documented in Tab.~\ref{tab:restapiendpoints} includes the endpoints used by front-end applications, such as the ones shown in Fig.~\ref{fig:frontEnds}. These endpoints support the creation, retrieval, and deletion of state machines.

\begin{table}[htb]
    \resizebox{\textwidth}{!}{%
    \begin{tabular}{l|l}
    \footnotesize
    Endpoint & Meaning \\ \hline \hline
    POST \dots/create & Creates new state machine as specified in request body \\ \hline
    GET \dots/all & Returns all state machines \\ \hline
    DELETE \dots/delete & Deletes state machine identified in request body \\ \hline \hline
    GET \dots/\{UUID\}/info & Returns name, description, and activity status \\ \hline
    POST \dots/\{UUID\}/respond & Returns the response to user uttereance in request body \\ \hline
    GET \dots/\{UUID\}/conversation & Returns the complete conversation so far \\ \hline
    PUT \dots/\{UUID\}/reset & Resets state machine to initial state
    \end{tabular}%
    }
    \caption{PROMISE REST API Endpoints (Subset)}
    \label{tab:restapiendpoints}
\end{table}

The \texttt{create} endpoint reads the specification of a state machine in JSON format, instantiates, and stores the resulting agent with the database. The \texttt{info}, \texttt{respond}, \texttt{conversation}, and \texttt{reset} endpoints provide the functionality to enable interaction front-ends such as the one shown on the left of Fig.~\ref{fig:frontEnds}. 

\begin{figure}[htb]
    \centering
    \includegraphics[width=\textwidth]{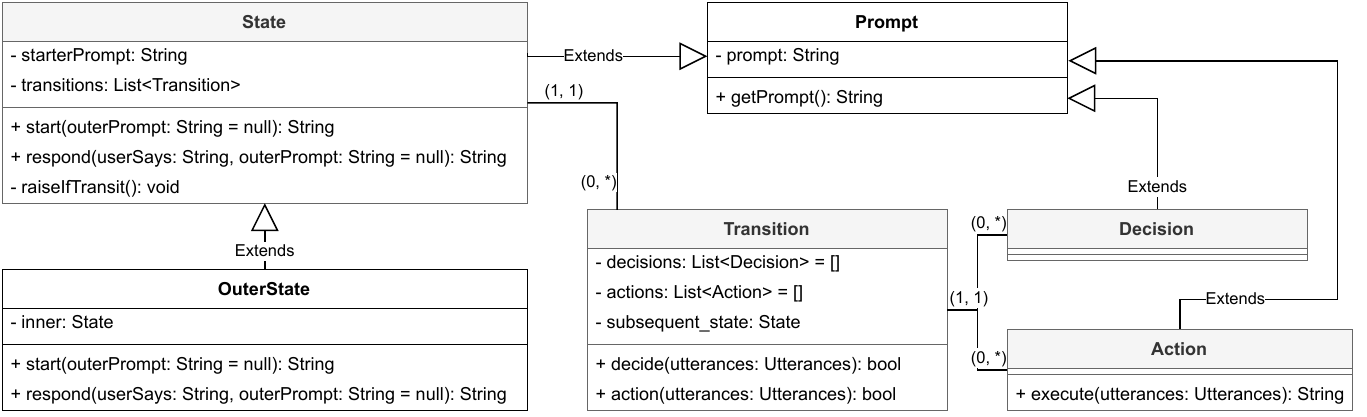}
    \caption{PROMISE Back-End Entity Classes (Subset)}
    \label{fig:frameworkClassDiagram}
\end{figure}

The REST API is implemented with a Spring Web \texttt{RestController}, which uses a Spring Data \texttt{JpaRepository} to access the database. The state machine concepts are implemented as JPA Entities and thus transparently persistent. The main entities are documented in the UML class diagram in Fig.~\ref{fig:frameworkClassDiagram}.

The \texttt{State} class implements the methods to start an interaction and respond to user utterances. A private \texttt{raise\-If\-Transit} method is called within a \texttt{respond} invocation to test if any transition should be followed. A \texttt{Tran\-sition} encapsulates transition decisions and action executions. An \texttt{Outer\-State} extends a state by an inner state, to which the requests for responses are forwarded. The \texttt{State}, \texttt{Decision}, and \texttt{Action} classes inherit from the superclass \texttt{Prompt}, which encapsulates the prompt and possible dynamic prompt augmentation behavior at runtime with the \texttt{getPrompt} method.

The \texttt{State} contains boolean configuration parameters not shown in the class diagram due to space limitations. For example, a state can be configured to open the interaction with a conversation starter as described in Sect.~\ref{sec:frameworkModel}. Additional configurations include whether a state is oblivious, in which case the repeated arrival in the state resets the previous interaction, or whether a state automatically checks for transitions without waiting for a user utterance.
\section{Validation of PROMISE}
\label{sec:validation}

To validate PROMISE, we demonstrate its capabilities for complex interactions beyond answering questions and following instructions. We present an example application to simulate patient interactions for physicians, providing them with immersive experiences that are monitored and guided to foster more compassionate communication. We focus on the language-based interaction and omit aspects of integrations with other system components due to space limitations.

Physicians converse with a fictitious patient simulated by the LM at one level of their interaction. At another level, the LM is instructed to act as a coach, observing the physician interacting with the patient. If a lack of compassion is detected, the coach gives advice. When the physician demonstrates compassion, the coach acknowledges the accomplishment, and the coaching session ends. 

\begin{figure}[htb]
    \centering
    \includegraphics[width=\textwidth]{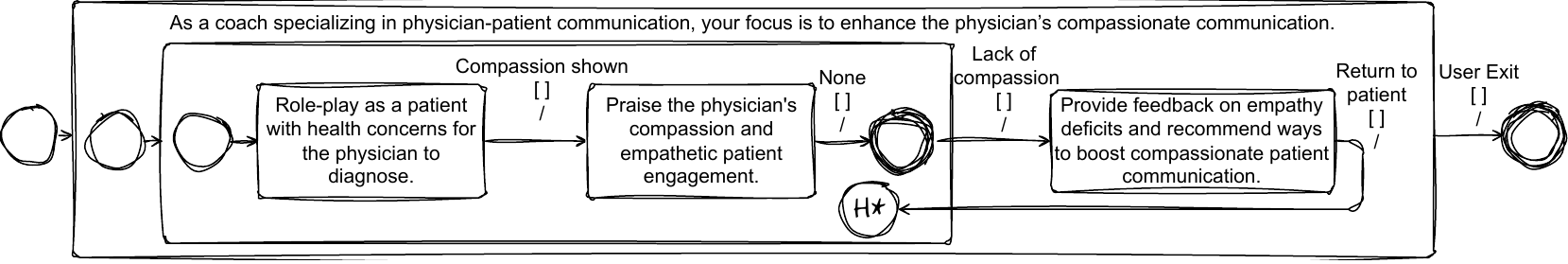}
    \caption{Model for Medical Consultation Coach}
    \label{fig:modelCompassionCoach}
\end{figure}

Figure~\ref{fig:modelCompassionCoach} shows a state machine describing this conversational behavior. The inner machine contains a state where the simulation takes place, from which a transition triggered by a demonstration of compassion points to a debriefing state where the physician's accomplishment is acknowledged. A surrounding outer state observes the conversation taking place in this inner machine.

The intervention is implemented with a transition from this outer state triggered by a lack of compassion. This transition points to a feedback state, calling out the lack of compassion and giving advice. This feedback state has a transition pointing back to the outer state containing the simulation. Notably, this transition points to a history node $H*$. This means that the interaction is automatically picked up at the state that was last active when the conversation left the outer state --- including the utterances collected in that state so far.

\begin{figure}[htb]
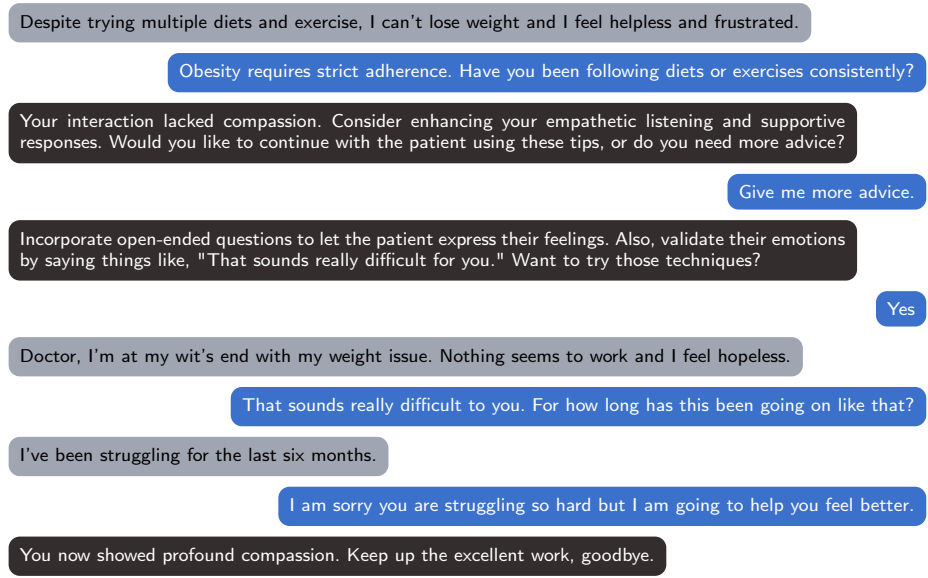

    \begin{leftbubbles}
        Despite trying multiple diets and exercise, I can't lose weight and I feel helpless and frustrated.
    \end{leftbubbles}
    \begin{rightbubbles}
        Obesity requires strict adherence. Have you been following diets or exercises consistently?
    \end{rightbubbles}
    \begin{leftbubblessecondary}
        Your interaction lacked compassion. Consider enhancing your empathetic listening and supportive responses. Would you like to continue with the patient using these tips, or do you need more advice?
    \end{leftbubblessecondary}
    \begin{rightbubbles}
        Give me more advice.
    \end{rightbubbles}
    \begin{leftbubblessecondary}
        Incorporate open-ended questions to let the patient express their feelings. Also, validate their emotions by saying things like, "That sounds really difficult for you." Want to try those techniques?
    \end{leftbubblessecondary}
    \begin{rightbubbles}
        Yes
    \end{rightbubbles}
    \begin{leftbubbles}
        Doctor, I'm at my wit's end with my weight issue. Nothing seems to work and I feel hopeless.
    \end{leftbubbles}
    \begin{rightbubbles}
        That sounds really difficult to you. For how long has this been going on like that?
    \end{rightbubbles}
    \begin{leftbubbles}
        I've been struggling for the last six months.
    \end{leftbubbles}
    \begin{rightbubbles}
        I am sorry you are struggling so hard but I am going to help you feel better.
    \end{rightbubbles}
    \begin{leftbubblessecondary}
        You now showed profound compassion. Keep up the excellent work, goodbye.
    \end{leftbubblessecondary}

    \caption{Multi-Layered Interaction: Simulation (Light Gray) \& Coaching (Dark)}
    \label{con:compassionCoach}
\end{figure}

Such a cyclic alignment of states and transitions, combined with nested states and history nodes, effectively supports the design of complex language-based interactions, including nested conversations. The conversation shown in Fig.~\ref{con:compassionCoach} is a simplified version of a coaching session demonstrating this multi-layered conversational behavior. All utterances of the fictitious patient are colored in light grey, while coaching utterances are colored in dark grey. The interaction maintains a seamless, single-conversation appearance to the physician due to the transparent transitions between different states.

Such a multi-layered conversation showcases one kind of support for complex interaction PROMISE provides. It is an example interaction that could not be provided reliably using sophisticated prompts only, and which would require much effort to code when using existing frameworks such as LangChain.
\section{Conclusion}
\label{sec:conclusion}

With the advent of more powerful language models, we are faced with inquiries about the feasibility and value of increasingly complex language-based interactions with information systems. Despite recent advances in prompting techniques and the availability of frameworks for developing LM applications, we note a need for more support for developing complex interactions using LMs.

We developed PROMISE, a framework supporting the design, implementation, and experimentation with complex interactions. PROMISE provides the means to represent such interactions using state modeling concepts. States specify their conversational behavior with a prompt, and transition triggers, guards, and actions may also be implemented using prompts. Complex interactions can thus be broken down into more specific prompts, increasing the LM behavior's predictability while harnessing their conversational capabilities. With the ability to hierarchically nest states and to put multiple transitions at different levels of the hierarchy, multi-layered interactions can be developed where the flow of interaction unfolds depending on decisions and actions taken within arbitrarily segmented interactions.

PROMISE overcomes the challenges of prompt engineering~\cite{chuSurveyChainThought2023} by segmenting complex interactions for improved control and predictability. It goes beyond frameworks such as LangChain~\cite{LangChainGitHub} by allowing explicit, nested segmentations and mappings of segments to actions and flows. With its semantically enriched state model, PROMISE advances past dialogue management~\cite{fastIrisConversationalAgent2018}, enabling dynamic prompt orchestration and efficient design in more diverse scenarios.

PROMISE has been effectively applied and tested in various projects, tackling challenges from improving health literacy in young adults with hearing loss to improving patient treatment adherence and tax morale. Notably, we observed that the use of state machine models not only facilitates the development process but fosters meaningful discussions of requirements and solutions with domain experts. This collaborative approach has been instrumental in refining and realizing these complex interactions more effectively.


\bibliographystyle{splncs04}
\bibliography{bibfile}

\end{document}